\documentclass{clv3}

\usepackage{hyperref}
\usepackage{xcolor}
\definecolor{darkblue}{rgb}{0, 0, 0.5}
\hypersetup{colorlinks=true,citecolor=darkblue, linkcolor=darkblue, urlcolor=darkblue}

\usepackage{booktabs}
\usepackage{float}
\usepackage{comment}
\usepackage{todonotes}

\usepackage{amsmath,amssymb,amsthm,bm}
\usepackage{tikz-dependency}
\usepackage{tikz-qtree}
\usepackage{tikz}
\usepackage{stfloats}
\usetikzlibrary{shapes,arrows,chains}
\usepackage{multirow}

\newcommand{\ccg}{{\tt CCG}}
\newcommand{\hpsg}{{\tt HPSG}}

\newcommand{\mrs}{{MRS}}
\newcommand{\dmrs}{{DMRS}}
\newcommand{\eds}{{EDS}}

\newcommand{\sdg}{{SDG}}
\newcommand{\amr}{{AMR}}

\DeclareMathOperator*{\argmax}{arg\,max}

\issue{}{}{}

\runningtitle{Knowledge-Intensive and Data-Intensive Semantic Parsers}

\runningauthor{Cao, Lin, Sun and Wan}

\begin{document}

%\title{A Comparative Analysis of Knowledge-Intensive and Data-Intensive Semantic Parsers}
%\title{Comparing Knowledge-intensive and data-intensive models for Elementary Dependency Structure Parsing}
\title{A Comparative Analysis of Knowledge-Intensive and Data-Intensive Semantic Parsers}

\author{Junjie Cao\thanks{Equal contribution. The work was done while the first three authors were at Peking University. Junjie Cao is now at Alibaba, and Zi Lin is now at Google.}}
\affil{Peking University\\Wangxuan Institute of Computer Technology\\\url{junjie.junjiecao@alibaba-inc.com}}

\author{Zi Lin$^*$}
\affil{Peking University\\Department of Chinese Language and Literature\\\url{lzi@google.com}}

\author{Weiwei Sun\thanks{Corresponding author. 
Now at the Department of Computer Science and Technology of University of Cambridge.}}
\affil{Peking University\\Wangxuan Institute of Computer Technology and Center for Chinese Linguistics\\\url{ws390@cam.ac.uk}}

\author{Xiaojun Wan}
\affil{Peking University\\Wangxuan Institute of Computer Technology\\\url{wanxiaojun@pku.edu.cn}}

\maketitle

\begin{abstract}
%In this work, we present a phenomenon-oriented comparative analysis of the two dominant approaches in task-independent semantic parsing: classic, knowledge-intensive and neural, data-intensive models. 
%To reflect state-of-the-art neural NLP technologies, a factorization-based ERS/EDS parser is introduced, which can produce semantic graphs much more accurately than previous data-driven parsers on English Resourse Semantics or Elementary Dependency Structure (ERS/EDS). 
%We conduct a suite of tests for different linguistic phenomena to analyze the grammatical competence of different parsers, where we show that, despite comparable performance overall, knowledge- and data-intensive models produce different types of errors, in a way that can be explained by their theoretical properties. 
In this work, we present a phenomenon-oriented comparative analysis of the two dominant approaches in English Resouce Semantic (ERS) parsing: classic, knowledge-intensive and neural, data-intensive models. 
To reflect state-of-the-art neural NLP technologies, a factorization-based parser is introduced, which can produce Elementary Dependency Structures much more accurately than previous data-driven parsers.
We conduct a suite of tests for different linguistic phenomena to analyze the grammatical competence of different parsers, where we show that, despite comparable performance overall, knowledge- and data-intensive models produce different types of errors, in a way that can be explained by their theoretical properties. 
This analysis is beneficial to in-depth evaluation of several representative parsing techniques and leads to new directions for parser development.
\end{abstract}

\section{Introduction}

Recent work in task-independent semantic parsing shifts from the knowledge-intensive approach to the data-intensive approach. 
Early attempts in semantic parsing leverage explicitly expressed symbolic rules in a deep grammar formalism, e.g., Combinatory Categorial Grammar  \citep[\ccg;][]{steedman1996surface,Steedman:2000}
and Head-driven Phrase Structure Grammar \citep[\hpsg;][]{hpsg94}, to model the syntactico-semantic composition process \citep{bos-EtAl:2004,pet}.
Then, statistical machine learning technologies, especially structured prediction models, are employed to enhance deep grammar-driven parsers \citep{CC:2007:CL,Miyao:2008:CL,zhang-oepen-carroll:2007:IWPT2007}.
Recently, various deep learning models together with vector-based embeddings induced from large-scale raw texts have been making advances significantly \citep{dozat-manning:2018:Short,shrgparser}.

This paper is concerned with comparing knowledge-intensive and data-intensive parsing models for 
English Resouce Semantics \citep[ERS;][]{ers,ersweb}, a comprehensive framework for in-depth linguistic analysis.
Figure \ref{fig:EDS} and \ref{fig:EDS_complicate} are two examples to illustrate the ERS representations.
Our comparison is based not only on the general evaluation metrics for semantic parsing, but also a fine-grained construction-focused evaluation that sheds light on the kinds of strengths each type of parser exhibits.
Characterizing such models may benefit parser development for not only ERS but also other frameworks, e.g. 
Groningen Meaning Bank \citep[GMB;][]{gmb2012,gmt} and 
Abstract Meaning Representation \citep[AMR;][]{amr}.
%In this work, we present a phenomenon-oriented comparative analysis of the knowledge-intensive and data-intensive 

To reflect the state-of-the-art deep learning technologies that are already available for data-intensive parsing, 
we design and implement a new factorization-based system for string-to-conceptual graph parsing \citep{graphbank}.
This parser learns to produce conceptual graphs for sentences from an annotated corpus and does not assume the existence of a grammar that explicitly defines syntactico-semantic patterns.
The core engine is scoring functions that use contextualized word and concept embeddings to discriminate good parses from bad for a given sentence, regardless of its semantic composition process.

To evaluate the effectiveness of the new parser, we conduct experiments on DeepBank \citep{flickinger2012deepbank}.
Our parser achieves an accuracy of 95.05\footnote{
  The results are obtained based on gold-standard tokenization.
}
for Elementary Dependency Structure \citep[\eds;][]{eds} in terms of \textsc{smatch},  which is 8.05 point improvement over the best transition-based model \citep{buys-blunsom:2017:Long} and  4.19 point improvement over the composition-based parser \citep{shrgparser}. 
We take it as a reflection that the models induced from large-scale data by neural networks have a strong coherence with linguistic knowledge.
Our parser has been re-implemented or extended by
two best-performing systems \citep{zhang-etal-2019-suda,chen-etal-2019-peking}
in the CoNLL 2019 Shared Task on Cross-Framework Meaning Representation Parsing \citep{conll-2019-shared}.

Despite the numerical improvements brought by neural networks, they have typically come at the cost of our understanding of the systems, i.e., it is still unclear as to what extent we can expect supervised training or pre-trained embeddings to induce the implicit linguistic knowledge and thus help semantic parsing. 
% In this paper, we take a step back and analyze the recent progress of semantic parsing through a comparative analysis of the knowledge-intensive and data-intensive paradigms. 
To answer this question, we utilize linguistically-informed datasets based on previous work \citep{bender-EtAl:2011:EMNLP} and create an extensive suite of other widely-discussed linguistic phenomena, covering a rich set of linguistic phenomena related to various lexical, phrasal and non-local dependency constructions. Based on the probing study, we find several non-obvious facts:
\begin{enumerate}
    \item The data-intensive parser is good at capturing local information at the lexical level even when the training data set is rather small.
    \item The data-intensive parser performs better on some peripheral phenomena but may suffer from data sparsity; % argument structure & phrasal construction
    \item The knowledge-intensive parser produces more coherent semantic structures, which may have a great impact on advanced natural language understanding tasks, such as textual inference.
    \item It is difficult for both parsers to find long-distance dependencies and their performance varies across phenomena. % non-local dependencies
\end{enumerate}
There is no apriori restriction that a data-intensive approach must remove all explicitly defined grammatical rules, or a knowledge-intensive approach cannot be augmented by data-based technologies.
Our comparative analysis appears highly relevant, in that these insights may be explored further to design new computational models with improved performance.\footnote{The code of our semantic parser, the test sets of linguistic phenomena as well as the evaluation tool can be found at \url{https://github.com/zi-lin/feds-parser} for research purposes.}

\section{Background} 
\subsection{Graph-Based Meaning Representations}

Considerable NLP research has been devoted to the transformation of natural language utterances into a desired linguistically-motivated semantic representation.
Such a representation can be understood as a class of discrete structures that describe lexical, syntactic, semantic, pragmatic as well as many other aspects of the phenomenon of human language.
In this domain, graph-based representations provide a light-weight yet effective way to encode rich semantic information of natural language sentences and have been receiving heightened attention in recent years \cite{graphbank}.
Popular frameworks under this umbrella includes Bi-lexical Semantic Dependency Graphs \citep[\sdg;][]{ccgdep,ergdep,sdp2014,sdp2015}, Abstract Meaning Representation \citep[\amr;][]{amr}, Graph-based Representations for ERS \citep{eds,slacker}, and Universal Conceptual Cognitive Annotation \citep[{\tt UCCA};][]{abend-rappoport:2013:ACL2013}.

\subsection{Parsing to Semantic Graphs: Knowledge-intensive v.s. Data-intensive}

Parsing to the graph-based representations has been extensively studied recently \cite{du-sun-wan:2015,tgrasscl,1ec2p,peng-thomson-smith:2017:Long,amrparsing14,artzi-lee-zettlemoyer:2015:EMNLP,peng-song-gildea:2015:CoNLL,buys-blunsom:2017:Long,hershcovich-abend-rappoport:2017:Long,konstas-EtAl:2017:Long,shrgparser}.
Work in this area can be divided into different types, 
according to how information about the mapping between natural language utterances and target graphs is formalized, acquired and utilized. 
In this paper, we are concerned with two dominant types of approaches in ERS parsing, which winned the `DM' and `EDS' sections of the CoNLL 2019 shared task \citep{conll-2019-shared}.

In the first type of approach, a semantic graph is derived according to a set of lexical and syntactico-semantic rules,  which extensively encode explicit linguistic knowledge. 
Usually, such rules are governed by a well-defined grammar formalism, e.g., \ccg, \hpsg, and Hyperedge Replacement Grammar, and exploit compositionality \citep{pet,bos-EtAl:2004,artzi-lee-zettlemoyer:2015:EMNLP,peng-song-gildea:2015:CoNLL,Gro:Lin:Fow:18,shrgparser}.
In this paper, we call it the \textbf{knowledge-intensive} approach.

The second type of approach explicitly models the target semantic structures.
It may associate each basic part with a target graph score, and casts parsing as the search for the graphs with the highest sum of partial scores \citep{amrparsing14,kuhtacl15,peng-thomson-smith:2017:Long}.
%To search for the highest-scoring semantic graphs involves a combinatorial optimization problem, which was usually resolved by dynamic programming.
%Strong neural encoders, such as LSTM \citep{lstm} and Transformer \citep{transformer}, are able to learn global information and 
The essential computational module in this architecture is the score function, which is usually induced based on moderate-sized annotated sentences.
%To evaluate the goodness of a semantic graph associated to an input sentence is to calculate the sum of \emph{local} scores assigned to those parts.
Various deep learning models together with vector-based encodings induced from large-scale raw texts have been making advances in shaping a score function significantly \citep{dozat-manning:2018:Short}.
This type of approach is also referred to as graph-based or factorization-based in the context of bi-lexical dependency parsing.
In this paper, we call it the \textbf{data-intensive} approach.

\begin{figure*}
%\centering
\caption{An example of \eds~graph. 
Some concepts are surface relations, meaning that they are related to a single lexical unit, e.g. {\tt \_the\_q} or {\tt \_introduce\_v\_to}, 
while others are abstract relations representing grammartical meanings, e.g. {\tt compound} (multiword expression), {\tt parg\_d} (passive) and {\tt loc\_nonsp} (temporal).
ERS corpus provides alignment between concept nodes and surface strings, e.g. {\tt <0:1>} that is associated to {\tt \_the\_q} indicates that this concept is signaled by the first word.}
\label{fig:EDS}
\vspace{1em}
\scalebox{0.7}{
\input{tikzed-graphs/eds_example_map_old_newstyle.tex}
}

  %$_0$ The $_1$ drug $_2$ was $_3$ introduced $_4$ in $_5$ West $_6$ Germany $_7$ this $_8$ year $_9$ . $_{10}$
  $_0$ The $_1$ drug $_2$ was $_3$ {\color{blue} introduced} $_4$ in $_5$ West $_6$ Germany $_7$ this $_8$ {\color{orange} year} $_9$ . $_{10}$
%\vspace{-0.3cm}
\end{figure*}

\begin{figure*}[!t]
%\centering
\caption{An example of \eds~graph to represent complicate phenomena like {\it right node raising} and {\it raising/control} constructions. 
% The graph shows {\tt house} and {\tt feed} share the same object {\tt poor} and 
  }
\label{fig:EDS_complicate}
\vspace{1em}
\scalebox{0.7}{
\input{tikzed-graphs/eds_example_complicate_newstyle.tex}
}
  %$_0$ The $_1$ drug $_2$ was $_3$ introduced $_4$ in $_5$ West $_6$ Germany $_7$ this $_8$ year $_9$ . $_{10}$
  
  $_0$ They $_1$ managed $_2$ to $_3$ house $_4$ and $_5$ feed $_6$ the $_7$ poor $_8$ . $_9$
  
%\vspace{-0.3cm}
\end{figure*}

\subsection{\textsc{delph-in} English Resource Semantics}
%\todo{More about ERS}
%\myworries{@Zi Lin: paraphrase and enrich this section. Say why we use ERS to study our problem: comparing knowledge- and data-intensive approaches.} 

In this paper, we take the representations from English Resource Semantics \citep[ERS;][]{ers}\footnote{\url{http://moin.delph-in.net/ErgSemantics}.} as our case study.
Compared to other meaning representations, ERS exhibits at least the following features: 
(1) ERS has a relatively high coverage for English text \citep{flickinger2010wikiwoods,flickinger2012deepbank,adolphs2008some}; 
(2) ERS has a strong transferability across difference domains \citep{copestake2000open,Iva:Oep:Dri:13}; 
(3) ERS has comparable and relatively high performance in terms of knowledge-intensive and data-intensive parsing technologies \citep{pet,zhang-oepen-carroll:2007:IWPT2007,shrgparser}.

ERS is the semantic annotation associated with English Resource Grammar \citep[ERG;][]{erg}, an open-source, domain-independent, linguistically precise and broad-coverage grammar of English, which encapsulates the linguistic knowledge required to produce many of the types of compositional meaning annotations. The ERG is an implementation of the grammatical theory of Head-driven Phrase Structure Grammar \citep[HPSG;][]{pollard1994head}.
ERG is a resource grammar that can be used for both parsing and generation. Development of the ERG begins in 1993, and after continuously evolving through a series of projects, it allows the grammatical analysis of most running text across domains and genres.

In the most recent stable release, viz. version `1214', the ERG contains 225 syntactic rules and 70 lexical rules for derivation and inflection. The hand-built lexicon of the ERG contains 39,000 lemmata, instantiating 975 leaf lexical types providing part-of-speech and valence constraints, which aims at providing complete coverage of function words and open-class words with `non-standard' syntactic properties (e.g., argument structure). The ERG also supports light-weight named entity recognition and an unknown word mechanism, allowing the grammar to derive full syntactico-semantic analyses for 85-95\% of all utterances in real corpora such as newspaper text, the English Wikipedia, or bio-medical academic paper \citep{flickinger2010wikiwoods,flickinger2012deepbank,adolphs2008some}. For more than 20 years of development, ERS has shown their advantages of explicit formalization and large scalability \citep{copestake2000open}. 

The Minimal Recursion Semantics \citep[\mrs;][]{mrs} is the associated semantic representation employed by the ERG. 
MRS is based on the first-order language family with generalized quantifiers.  
ERS can also be expressed as other semantic graphs, including \sdg~\citep{ergdep}, \eds~\citep{eds} and  Dependency-based Minimal Recursion Semantics \citep[\dmrs;][]{slacker}. 
In this paper, we illustrate our models using the \eds~format\footnote{\url{http://moin.delph-in.net/EdsTop}}.
% recover cjj
% recover done

\begin{figure}[ht]
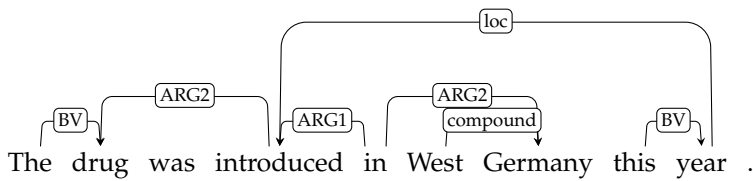

%\centering
\caption{The standard \sdg~that is converted from the \eds~in Figure \ref{fig:EDS}.}
\vspace{1em}
\scalebox{1}{
    \begin{dependency}[edge above, edge slant=0.15ex, edge unit distance=2ex]
      \begin{deptext}[column sep = 0.1cm]
        The \& drug \& was \& introduced \& in \& West \& Germany \& this \& year \&.\\
      \end{deptext}
      \depedge{1}{2}{BV}
      \depedge{4}{2}{ARG2}
      \depedge{5}{4}{ARG1}
      \depedge[edge start x offset=8pt]{9}{4}{loc}
      \depedge{5}{7}{ARG2}
      \depedge{6}{7}{compound}    
      \depedge{8}{9}{BV}
      
    \end{dependency}
}

\label{fig:sdg}
\end{figure}

The graphs in Figure \ref{fig:EDS} and  \ref{fig:EDS_complicate} are examples of \eds~graph.
Figure \ref{fig:EDS_complicate} further shows \eds~does have the ability to represent more complicated linguistic phenomena such as the {\it right node raising} and {\it raising/control} constructions\footnote{{\it Right node raising} often involves coordination where they share the same component (e.g., the subject {\it they} here for the predicates {\it house} and {\it feed}); the {\it raising/control} construction refers to raising and control verbs that select for an infinitival VP complement and stipulate that another of their arguments (subject or direct object in the example) is identified with the unrealized subject position of the infinitival VP. For further details, see \citet{bender-EtAl:2011:EMNLP}.}
The semantic structure is a directed graph where nodes labeled with semantic predicates/relations related to a constituent of the sentence, and arcs are labeled with semantic arguments roles.
By linking concepts with lexical units, this \eds~graph can be reduced to an \sdg, as shown in Figure \ref{fig:sdg}. 
In these forms, {\it relation} is the predicate name of an Elementary Predication from the \mrs, and {\it role} is an argument label such as {\it ARG1}. 
%\citeauthor{ergdep}'s transformation can be applied in our creation of evaluation set for different linguistic phenomena in Table \ref{tb:linguistic phenomena}.

\subsection{Analyzing Neural Networks for NLP}

%\myworries{@Zi Lin: revise this section} 

What are the representations that the neural network learns and how can we explore that? 
Concerns of this question have led to the interpretability of the system being an active area of research.
Related work tries to answer these questions by: (1) investigating specific components of the architectures \citep{karpathy2015visualizing,radford2017learning,qian2016analyzing,bau2018identifying}, (2) testing models on specific tasks, including part-of-speech tagging \citep{blevins2018deep,belinkov2017evaluating,shi2016does}, semantic role labeling \citep{tenney2018you}, word sense disambiguation \citep{peters2018deep}, coreference \citep{peters2018dissecting}, etc., and (3) building linguistically-informed dataset for evaluation \citep{linzen2016assessing,isabelle2017challenge,burchardt2017linguistic,sennrich2017grammatical,isabelle2018challenge,warstadt2018neural,wang2018glue}.

In this paper, we try to probe this question by applying the models built on the state-of-the-art technologies to the string-to-conceptual graph parsing task, and utilizing linguistically-informed datasets based on previous work \citep{bender-EtAl:2011:EMNLP} and our own creation.

\section{A Knowledge-Intensive Parser}
There are two main existing knowledge-intensive parsers with unification-based grammars for the English Resource Grammar: 
      the PET system\footnote{\url{http://moin.delph-in.net/PetTop}} \citep{pet,Zha:Oep:Car:07} and the ACE system\footnote{\url{http://sweaglesw.org/linguistics/ace/}}.
PET is an efficient open-source parser for unification grammars. Coupled with ERG, it can produce HPSG-style  syntactico-semantic derivations and \mrs-style~semantic representations in logic forms.
Similar to PET, ACE is another industrial strength implementation of the typed feature structure formalism.
Note that the key algorithms implemented by PET and ACE are the same.
We choose to use ACE in this work given the fact that, comparing to PET's parsing performance, ACE is faster in certain common configurations.
Coupled with ERG, it serves as a valid companion to study our problem: comparing knowledge- and data-intensive approaches.

\section{A Data-Intensive Parser}

To empirically analyze data-intensive parsing technologies,
we design, implement and evaluate a new target structure--centric parser for ERS graphs, trained on \emph{(string, graph)} pairs without explicitly incorporating linguistic knowledge.
The string-to-graph parsing is formulated as a problem of finding the Maximum Subgraph for a graph class $\mathcal{G}$ of a sentence $s=l_1,\ldots,l_m$:
Given a graph $G=(V,A)$ related to $s$, the goal is to search for a subset $A'\subseteq A$ with maximum total score such that
the induced subgraph $G'=(V,A')$ belongs to $\mathcal{G}$.
Formally, we have the following optimization problem:
\[
  \arg\max_{G^*\in\mathcal{G}(s,G)}\sum_{p\in\textsc{factorize}(G^*)}\textsc{score}_{\text{part}}(s,p)
\]
where $\mathcal{G}(s,G)$ denotes the set of all graphs belong to $\mathcal{G}$ and compatible with $s$ and $G$.
This view matches a classic solution to the structured prediction which captures elemental and structural information through part-wise factorization.
To evaluate the goodness of a semantic graph is to calculate the sum of \emph{local} scores assigned to those parts.

In the literature of bi-lexical dependency parsing, the above architecture is also widely referred to as factorization-based.
as such a parser factors all valid structures for a given sentence into smaller units, which can be scored somewhat independently.

For string-to-graph parsing, we consider two basic factors, i.e., single concepts and single dependencies.
Formally, we use the following objective function:
\[
  \sum_{n\in\textsc{node}(G)}\textsc{sc}_{\text{n}}(s,n)+\sum_{(p,a)\in\textsc{arc}(G)}\textsc{sc}_{\text{a}}(s,p,a)
\]

%\subsection{A Two-step Architecture}
Our parser adopts a two-step architecture to produce \eds~graphs:
(1) it identifies the concept nodes based on contextualized word embeddings by solving a simplified optimization problem, viz. $\max\sum_{n\in\textsc{node}(G)}\textsc{sc}_{\text{n}}(s,n)$;
(2) it identifies the dependencies between concepts based on concept embeddings by solving another optimization problem, viz. $\max\sum_{(p,a)\in\textsc{arc}(G)}\textsc{sc}_{\text{a}}(s,p,a)$.
Particularly, our architecture is a pipeline: single best prediction of the first step is utilized as the input for the second step.

\subsection {Concept Identification}

Usually, the nodes in a conceptual graph have a strong correspondence to surface lexical units, viz. tokens, in a sentence.
Take the graph in Figure \ref{fig:EDS} for example, the generalized quantifier {\tt \_the\_q} corresponds to \emph{the} and the property concept {\tt \_drug\_n\_1} corresponds to \emph{drug}.
Because the concepts are highly \emph{lexicalized}, it is reasonable to employ a sequence labeler to predict concepts triggered by tokens.

Nodes may be aligned with arbitrary parts of the sentence, including sub-token or multi-token sequences, which affords
more flexibility in the representation of meaning contributed by derivational morphemes (e.g., {\tt parg\_d} that indicates a passive construction) or phrasal constructions (e.g., {\tt compound\_name} that indicates a multiword expression).
To handle these types of concepts by a word-based sequence labeler, we align them to words based on their span information and a small set of heuristic rules.
Take Figure \ref{fig:EDS_mapping} for example, we align {\tt parg\_d} to the word where \emph{-ed} is attached to, and {\tt compound\_name} to the first word of the compound. 

\begin{figure*}[htb]
  \caption{An example for illustrating concept identification.
  The ``\textsc{n}'' row presents the results of lexicalization.
  The ``\textsc{s}'' row presents the gold tags assigned to tokens which are utilized to train a sequence labeling based concept identifier.
  }
  \label{fig:EDS_mapping}
  %\centering
  \vspace{1em}
  \scalebox{0.63}{
  \begin{tabular}{l|c|c|c|c|c|c|c|c|c}
    \toprule
     & The  & drug  & was  & introduced & in & West & Germany & this & year \\
    \midrule
    \multirow{2}{*}{\textsc{n}} & {\tt \_the\_q} & {\tt \_drug\_n\_1} & $\emptyset$ & {\tt \_introduce\_v\_to} & {\tt \_in\_p} & {\tt named}          & {\tt named}     & {\tt \_this\_q\_dem} & {\tt \_year\_n\_1} \\
    &                &                    &             & {\tt parg\_d}            &               & {\tt compound\_name} & {\tt proper\_q} &                      & {\tt loc\_nonsp} \\
    &                &                    &             &             &               & {\tt proper\_q} &  &                      &  \\
    \midrule
    \multirow{2}{*}{\textsc{s}} & {\tt *\_q} & {\tt *\_n\_1} & $\emptyset$ & {\tt *\_v\_to} & {\tt *\_p} & {\tt named}          & {\tt named}     & {\tt *\_q\_dem} & {\tt *\_n\_1} \\
    &                &                    &             & {\tt parg\_d}            &               & {\tt compound\_name} & {\tt proper\_q} &                      & {\tt loc\_nonsp} \\
    &                &                    &             &             &               & {\tt proper\_q} &  &                      &  \\
    \bottomrule
  \end{tabular}}
\end{figure*}

The concept predicate may contain the lexical part aligning to the surface predicate, which leads to a serious data sparseness problem for training. %such as {\it drug} of {\tt\_drug\_n\_1} in Figure \ref{fig:EDS}.
To deal with this problem, we {\it delexicalize} lexical predicates as described in \citet{buys-blunsom:2017:Long}:
replacing the lemma part by a placeholder ``{\tt *}". 
Figure \ref{fig:EDS_mapping} shows a complete example. 
In summary, the concept identification problem is formulated as a word tagging problem: 
\[
  \sum_{n\in\textsc{node}(G)}\textsc{sc}_{\text{n}}(s,n) \approx \sum_{1\le i\le m}\max_{st_i\in \mathcal{ST}}\textsc{sc}_{\text{st}}(s,i,st_i)
\]

Our parser applies a neural sequence labeling model to predicts concepts.
In particular, a BiLSTM model is utilized to capture words' contextual information and another softmax layer for classification.
Usually, words and POS-tags are needed to be transformed into continuous and dense representation in neural models. 
Inspired by \citet{costajussa-fonollosa:2016:P16-2}, we use word embedding of two granularities in our model: 
character-based and word-based, for low frequency and high-frequency words (the words appear more than $k$ times in the training data) respectively. 
A character-based model can capture rich affix information of low-frequency words for better word representations.
The word-based embedding uses a common lookup-table mechanism. The character-based word embedding $\bm{w}_i$ is implemented by extracting features with bidirectional LSTM from character embeddings $\bm{c}_1, \ldots,\bm{c}_n$:

Contextualized representation models such as CoVe \citep{mccann2017learned}, ELMo \citep{peters2018deep}, OpenAI GPT \citep{radford2018improving} and BERT \citep{devlin2018bert} have recently achieved the state-of-the-art results on downstream NLP models across many domains. 
In this paper, we use pretrained ELMo model and learn a weighted average of all ELMo layers for our embedding $\bm{e}_i$ to capture richer contextual information.
The concatenation of word embedding $\bm{w}_i$, ELMo embedding and POS-tag embedding $\bm{t}_i$ of each word in a specific sentence is used as the input of bi-LSTMs to extract context-related feature vectors $r_i$ for each position $i$. 
Finally we use $\bm{r}_i$ as input of a softmax layer to get the probability $\textsc{sc}_{\text{st}}(s,i,st_i)$.

\begin{align*}
\bm{a}_i &= \bm{w}_i \oplus \bm{e}_i \oplus \bm{t}_i \\
\bm{r}_1:\bm{r}_m &= \text{BiLSTM}(\bm{a}_1:\bm{a}_m)\\
  \textsc{sc}_{\text{st}}(s, i, st_i) &= \text{softmax}(\bm{r}_i) \\
\end{align*}

\subsection{Dependency Identification}
Given a set of concept nodes $N$ which are predicted by our concept identifier, the semantic dependency identification problem is formalized as the following optimization problem:
 \[
\hat{G} = \argmax_{G\in\mathcal{G}(N)}\sum_{(p,a) \in \textsc{arc}(G)}{\textsc{sc}_{\text{a}}(s,N,p,a)}
\]
where $\mathcal{G}(N)$ denotes the set of all possible graphs that take $N$ as their vertex set.
Following the factorization principle, we measure a graph using a sum of local scores.

In order to effectively learn a local score function, viz. $\textsc{sc}_a$, we represent concepts with the concatenation of two embeddings: textual and conceptual embeddings.
\begin{align*}
\bm{c}_i &= \bm{r}_i \oplus \bm{n}_i
\end{align*}
To represent two concept nodes' textual information,    
we use stacked BiLSTMs that are similar to the proposed structure of our concept identifier to get $\bm{r}_i$.

Besides contextual information, we also need to transform a concept into a dense vector $\bm{n}_i$.
Similar to word embedding and POS-tag embedding, we also use a common lookup-table mechanism and let our parser automatically induce conceptual embeddings from semantic annotations.

We calculate scores for all directional arcs between two concepts in the graph, which can be scored with a non-linear transformation from the two feature vectors of each concept pair:
\[
\textsc{$\textsc{sc}_{{a}}$}(s,N,p,a) = \bm{W}_2 \cdot \mathrm{\delta}(\bm{W}_{1} \cdot (\bm{c}_p \oplus \bm{c}_a) + \bm{b}) 
\]

Similar to unlabeled arcs, we also use MLP to get each arc's scores for all labels, and select the max one as its label.

For training, we use a margin-based approach to compute loss from the gold graph $G^*$ and the best prediction $\hat{G}$ under the current model and decoder. 
We define the {\it loss} term as:
$$
\max(0, \Delta(G^*, \hat{G}) - \textsc{Score}(G^*) + \textsc{Score}(\hat{G}))
$$
The margin objective $\Delta$ measures the similarity between the gold graph $G^*$ and the prediction $\hat{G}$. Following \citet{peng-thomson-smith:2017:Long}'s approach, we define $\Delta$ as weighted Hamming to trade-off between precision and recall.

Inspired by the \textbf{maximum spanning connected subgraph} algorithm proposed by \citet{amrparsing14}, we also consider using an additional constraint to restrict the generated graph to be connected.
The algorithm is simple and effective:
generating a maximum spanning tree (MST) firstly, and then adding all arcs with positive local scores. 
During the training, our dependency identifier ignores this constraint.

\subsection{Evaluation}
We conduct experiments on DeepBank v1.1 that corresponds to ERG version 1214, and adopt the {\it standard} data split.
The pyDelphin\footnote{\footnotesize{\url{www.github.com/delph-in/pydelphin}}} library and the jigsaw tool\footnote{\footnotesize{\url{www.coli.uni-saarland.de/~yzhang/files/jigsaw.jar}}} are leveraged to extract \eds~graphs and to separate punctuations from their attached words respectively.
The TensorFlow {\it ELMo} model\footnote{\footnotesize{\url{www.github.com/allenai/bilm-tf}}} is trained on the 1B Word Benchmark for pre-trained feature, and we use the same pre-trained word embedding introduced in \citet{yoav2016}.
DyNet2.0\footnote{\footnotesize{\url{www.github.com/clab/dynet}}} is utilized to implement the neural models. The automatic batch technique \citep{autobatch} in DyNet is applied to perform mini-batch gradient descent training, where the batch size equals to 32. 

Different models are tested to explore the contribution of BiLSTM and ELMo, including (1) {\it ELMo*} using BiLSTM and ELMo features, (2) {\it ELMo} using only ELMo features and softmax layer for classification, (3) {\it W2V} using BiLSTM and word2vec features \citep{mikolov2013distributed} and (4) {\it Random} using BiLSTM and random embedding initialization.
In all these experiments, we only learn a weighted average of all biLM layers but froze other parameters in ELMo.

\subsubsection{Results on Concept Identification}

Since we predict concepts by composing them together as a word tag, there are two strategies for evaluating concept identification: 
the accuracy of tag (viz. concept set) prediction and concept (decomposed from tags) prediction. 
For the former, we take ``$\emptyset$'' as a unique tag and compare each word's predicted tag as a whole part; for the latter, we ignore the empty concepts, such as {\it was} in Figure \ref{fig:EDS}. 
We can see that the {\it ELMo*} model performs better than the others. Empirically speaking, the numeric performance of concept prediction is better than the tag prediction. 
The results are illustrated in Table \ref{table:Concept_predication}. 
\begin{table}[ht]
  %\centering
  \caption{\label{table:Concept_predication}
  Accuracy of concept identification on development data.
  }
  \scalebox{1}{ 
  \begin{tabular}{l|c|ccc}
    \toprule
        &  Tag  &  \multicolumn{3}{c}{Concept} \\
        \cmidrule{2-5}
        &  Accuracy  &  Precision  & Recall & F-Score  \\

    \midrule
    Random & 92.74  & 95.13  & 94.60  & 94.87  \\ 
    W2V & 94.51  & 96.68  & 96.11  & 96.39  \\ 
    ELMo & 92.34  & 95.73  & 95.16  & 95.45  \\ 
    ELMo* & 95.38  & 97.31  & 96.77  & 97.04  \\ 
    \bottomrule  
  \end{tabular}
  }
\end{table}

\subsubsection{Results on Dependency Identification}
In dependency identification step, we train the parsing model on sentences with golden concepts and alignment. 
Both unlabeled and labeled results are reported.
Since golden concepts are used, the accuracy will obviously be much higher than the total system with predicted concepts.
Nevertheless, the numbers here serve as a good reflection of the goodness of our models.
We can see that %the {\it ELMo*} model still performs the best, but the {\it ELMo} model is much lower than the others, indicating that BiLSTM layers are much more important for dependency identification. 
Table \ref{table:Arc_predication} shows the results.
The measure for comparing two dependency graphs is the precision/recall of concept tokens which are defined as $\langle c_h,c_d,l\rangle$ tuples,
where $c_h$ is the functor concept, $c_d$ is the dependent concept and $l$ is their dependency relation.
Labeled precision/recall (LP/LR) is the ratio of tuples correctly identified by the automatic generator,
  while unlabeled precision/recall (UP/UR) is the ratio regardless of $l$.
F-score (LF/UF) is a harmonic mean of precision and recall.
\begin{table}[ht]
  %\centering
  \caption{Accuracy of dependency identification on development data.}
  \scalebox{1}{
   \begin{tabular}{l|ccc|ccc}
    \toprule
    Model  &  UP  &  UR & UF & LP & LR & LF  \\\midrule
    Random & 94.47  & 95.95  & 95.21  & 94.25  & 95.57  & 94.98 \\
    W2V  & 94.91  & 96.3  & 95.6  & 94.72  & 96.12  & 95.42 \\
    ELMo & 88.97  & 92.95  & 90.92  & 88.44  & 92.40  & 90.38 \\
    ELMo* & 96.00  & 96.99  & 96.49  & 95.80  & 96.79  & 96.29 \\
    \bottomrule  
  \end{tabular}}
  \label{table:Arc_predication}
\end{table}

\subsubsection{Results for Graph Identification}

As for the overall evaluation, we report parsing accuracy in terms of \textsc{smatch} \citep{cai2013smatch} that considers both nodes and relations, which was initially used for evaluating AMR parsers.
The Smatch metric \citep{cai2013smatch}, proposed for evaluating AMR graphs, also measures graph overlap, but does not rely on sentence alignments to determine the correspondences between graph nodes. 
\textsc{smatch} is computed by performing inference over graph alignments to estimate the maximum F-score obtainable from a one-to-one matching between the predicted and gold graph nodes.
Different from EDM \citep{Dri:Oep:11}, We only use each node's predicate but ignore the span information while aligning the two nodes.
But the results of these two evaluations are positively related. 

Considering the difference between the AMR graph and the \eds~graph, we implement our own tool for the disconnected graph, and calculate the scores in Table \ref{table:smatch_dev}.
The {\it ELMo*}'s concept and arc score are obviously higher than the others, while {\it ELMo}'s arc prediction yields the lowest \textsc{smatch} score.
\begin{table}[ht]
  %\centering
    \caption{\label{table:smatch_dev} Accuracy of the whole graphs on the development data. 
  {\it Concept} and {\it Arc} in the table header are the F-Score of concept and arc mapping for the highest \textsc{smatch} score. 
{\it Smatch} is the Smatch score of each model.
}
  \scalebox{1}{
  \begin{tabular}{l|ccc}
    \toprule
    Model &  Concept & Arc & \textsc{smatch}  \\\midrule
    Random & 94.69  & 91.25  & 92.95 \\ 
    W2V & 96.07  & 92.41  & 94.22 \\ 
    ELMo & 95.06  & 86.75  & 90.82 \\ 
    ELMo* & 96.71  & 93.86  & 95.27 \\ 
    \bottomrule
  \end{tabular}}

\end{table}

We compare our system with the ERG-guided ACE parser, the data-driven parser introduced in \citet{buys-blunsom:2017:Long} and composition-based parser in \citet{shrgparser} on the test data.
As ACE parser fails to parse some sentences (more than 1\%)\footnote{Note that the DeepBank data already removes a considerable portion (c.a. 11\%) of sentences.}), 
the outputs of the whole data and the successfully parsed part are evaluated separately.
For the other parsers, the results on the whole data and those ACE parsed data are very similar (less than 0.05\% lower), so we show the results on the whole data for brevity.
The numbers of ACE and \citet{buys-blunsom:2017:Long}'s are different from the results as they reported due to the different \textsc{smatch} evaluation tools.
Our {\it ELMo*} model achieves an accuracy of 95.05, which is significantly better than existing parsers,
demonstrating the effectiveness of this parsing architecture.

\begin{table}[ht]
\caption{\label{table:smatch_tst} Accuracy (\textsc{smatch}) on the test data.
  ACE$_1$ is evaluated on the whole data set: sentences that do not receive results are taken as empty graphs. 
  ACE$_2$ is evaluated on the successfully parsed data only.}
  %\centering
  \scalebox{1}{
  \begin{tabular}{l|ccc}
    \toprule
     Model    &  Node & Arc & \textsc{smatch} \\
    \midrule
    ACE$_1$ & 95.51  & 91.90 & 93.69 \\
    ACE$_2$ & 96.42  & 92.84 & 94.61 \\
    \citet{buys-blunsom:2017:Long}  & 89.06  & 84.96 & 87.00 \\
   \midrule
    %\citet{shrgparser} & 94.51  & 87.29  & 90.86 \\
    \citet{shrgparser} & 94.47  & 88.44  & 91.42 \\
   \midrule
    W2V & 95.65  & 91.97  & 93.78 \\ 
    ELMo & 94.74  & 86.64  & 90.60 \\ 
    ELMo* & 96.42  & 93.73  & 95.05 \\ 
    \bottomrule
  \end{tabular}}
\end{table}

\section{Linguistic Phenomena in Question}
\label{sec:linguistic phenomena}
Most benchmark datasets in NLP are drawn from text corpora, reflecting a natural frequency distribution of language phenomena. 
Such datasets are usually insufficient for evaluating and analyzing neural models in many advanced NLP tasks, since they may fail to capture a wide range of complex and low-frequency phenomena \citep{kuhnle2018deep, belinkov2018analysis}. Therefore, an extensive suite of unit-tests should be considered to evaluate models on their ability to handle specific linguistic phenomena \citep{lin2019parsing}.

% \begin{table*}[ht]
\begin{table*}
\caption{Definitions and examples of the linguistic phenomena for evaluation.}
\label{tb:linguistic phenomena}
\vspace{1em}
\scalebox{0.72}{
\begin{tabular}{m{3.3cm}|m{3.8cm}|m{11.2cm}}
\toprule
\multicolumn{1}{c|}{Definition} & \multicolumn{1}{c|}{Head} & \multicolumn{1}{c}{Examples} \\ \midrule

\begin{tabular}[c]{@{}l@{}}{\tt comp}: compound/\\named entity\end{tabular}& \begin{tabular}[c]{@{}l@{}}head word in compound\\or last name\end{tabular} &
%\begin{tabular}[c]{@{}l@{}}{\tt comp}: compound/\\named entity\end{tabular}& \begin{tabular}[c]{@{}l@{}}head word in compound\\or last name\end{tabular} &
\begin{dependency}[edge above, edge slant=0.15ex, edge unit distance=0.8ex]
  \begin{deptext}[column sep=1ex]
    Donald \& Trump \& withdrew \& his \& \$7.54 billion \& offer.\\
  \end{deptext}
  \depedge{2}{1}{compound}
  \depedge{6}{5}{compound}
\end{dependency} \\ \midrule

\begin{tabular}[c]{@{}l@{}}{\tt as}: basic predicate-\\argument structures\end{tabular}& \begin{tabular}[c]{@{}l@{}}core predicate\end{tabular} &
\begin{dependency}[edge above, edge slant=0.15ex, edge unit distance=0.4ex]
  \begin{deptext}[column sep=1ex]
    Mike \& gave \& them \& to \& a \& new \& bureaucracy.\\
  \end{deptext}
  \depedge{2}{1}{ARG1}
  \depedge{2}{3}{ARG2}
  \depedge{2}{7}{ARG3}
\end{dependency} \\ \midrule

 \begin{tabular}[c]{@{}l@{}}{\tt ditr}: ditransitive\\construction\end{tabular}& \begin{tabular}[c]{@{}l@{}}core predicate\end{tabular} &
\begin{dependency}[edge above, edge slant=0.15ex, edge unit distance=0.6ex]
  \begin{deptext}[column sep=1ex]
    Sally \& baked \& her \& sister \& a \& cake.\\
  \end{deptext}
  \depedge{2}{1}{ARG1}
  \depedge{2}{6}{ARG2}
  \depedge{2}{4}{ARG3}
\end{dependency} \\ \midrule

\begin{tabular}[c]{@{}l@{}}{\tt causemo}: cause\\motion construction\end{tabular}& \begin{tabular}[c]{@{}l@{}}core predicate\end{tabular} &
\begin{dependency}[edge above, edge slant=0.15ex, edge unit distance=0.6ex]
  \begin{deptext}[column sep=1ex]
    The \& audience \& laughed \& Bob \& \& off \& the stage.\\
  \end{deptext}
  \depedge{3}{1}{ARG1}
  \depedge{3}{4}{ARG2}
  \depedge{3}{6}{ARG3}
\end{dependency} \\ \midrule

\begin{tabular}[c]{@{}l@{}}{\tt way}: way construction\end{tabular}& \begin{tabular}[c]{@{}l@{}}core predicate\end{tabular} &
\begin{dependency}[edge above, edge slant=0.15ex, edge unit distance=0.6ex]
  \begin{deptext}[column sep=1ex]
    Frank \& dug \& his \& way \& \& out of \& prison.\\
  \end{deptext}
  \depedge{2}{1}{ARG1}
  \depedge{2}{4}{ARG2}
  \depedge{2}{6}{ARG3}
\end{dependency} \\ \midrule

\begin{tabular}[c]{@{}l@{}}{\tt passive}: passive\\verb construction\end{tabular}& \begin{tabular}[c]{@{}l@{}}Passive verb\end{tabular} &
\begin{dependency}[edge above, edge slant=0.15ex, edge unit distance=0.6ex]
  \begin{deptext}[column sep=1ex]
    The \& paper \& was \& accepted \& by \& the reviewer.\\
  \end{deptext}
  \depedge{4}{2}{ARG2}
  \depedge{4}{6}{ARG1}
\end{dependency} \\ \midrule

\begin{tabular}[c]{@{}l@{}}{\tt vpart}: verb-particle\\constructions\end{tabular}& \begin{tabular}[c]{@{}l@{}}({\color{blue}B}) verb+particle\end{tabular} &
\begin{dependency}[edge above, edge slant=0.15ex, edge unit distance=0.4ex]
  \begin{deptext}[column sep=1ex]
    The \& pass \& helped \& set up \& Donny's \& two \& companies.\\
  \end{deptext}
  \depedge{4}{7}{ARG2}
\end{dependency} \\ \midrule

\begin{tabular}[c]{@{}l@{}}{\tt itexpl}: expletive {\it it}\end{tabular} & \begin{tabular}[c]{@{}l@{}}{\it it}-subject taking verb\end{tabular} &
\begin{dependency}[edge above, edge slant=0.15ex, edge unit distance=0.4ex]
  \begin{deptext}[column sep=1ex]
    It \& is \& suggested \& that \& the \& flight \& was \& canceled.\\
  \end{deptext}
  \depedge{3}{1}{ARG2}
\end{dependency} \\ \midrule
%{\it It} {\bf is suggested} that the flight crew were drunk\\ \midrule

\begin{tabular}[c]{@{}l@{}}{\tt ned}: adj/Noun2+\\Noun1-{\it ed}\end{tabular}& \begin{tabular}[c]{@{}l@{}}({\color{red}A}) head noun\\({\color{blue}B}) Noun1-{\it ed}\end{tabular} &
\begin{dependency}[edge above, edge slant=0.15ex, edge unit distance=1ex]
  \begin{deptext}[column sep=1ex]
    The \& light \& colored \& glazes \& have \& softening \& effects.\\
  \end{deptext}
  \depedge{3}{2}{MOD({\color{red}A})}
  \depedge{4}{3}{MOD({\color{blue}B})}
\end{dependency} \\ \midrule
%{\it Light} {\it  \textbf{colored}} {\bf glazes} also have softening effects.\\ \midrule

\begin{tabular}[c]{@{}l@{}}{\tt argadj}: interleaved\\arg/adj\end{tabular}& \begin{tabular}[c]{@{}l@{}}({\color{red}A}) selecting verb\\({\color{blue}B}) selecting verb\end{tabular} &
\begin{dependency}[edge above, edge slant=0.15ex, edge unit distance=0.4ex]
  \begin{deptext}[column sep=1ex]
    The \& story \& shows, \& through \& flashbacks, \& the \& different \& histories.\\
  \end{deptext}
  \depedge{3}{4}{MOD({\color{red}A})}
  \depedge{3}{8}{ARG2({\color{blue}B})}
\end{dependency} \\ \midrule
% The story {\bf shows}, {\it through} flashbacks, the different {\it histories} of the characters. \\ \midrule

\begin{tabular}[c]{@{}l@{}}{\tt barerel}: bare\\relatives ({\it that}-less)\end{tabular}& \begin{tabular}[c]{@{}l@{}}({\color{blue}B}) grapped predicate\\in relative\end{tabular} &
\begin{dependency}[edge above, edge slant=0.15ex, edge unit distance=0.4ex]
  \begin{deptext}[column sep=1ex]
    They \& took \& over \& the \& lead \& (that) \& brooklyn \& has \& held.\\
  \end{deptext}
  \depedge{9}{5}{ARG2}
\end{dependency} \\ \midrule

\begin{tabular}[c]{@{}l@{}}{\tt tough}: tough\\adjectives\end{tabular} & \begin{tabular}[c]{@{}l@{}}({\color{red}A}) tough adjective\\({\color{blue}B}) grapped predicate\\in {\it to}-VP\end{tabular} &
\begin{dependency}[edge above, edge slant=0.15ex, edge unit distance=0.4ex]
  \begin{deptext}[column sep=1ex]
    Original \& copies \& are \& very \& hard \& to \& find.\\
  \end{deptext}
  \depedge{7}{2}{ARG2({\color{blue}B})}
  \depedge{5}{7}{ARG1({\color{red}A})}
\end{dependency} \\ \midrule
%Original {\it copies} are very hard to {\bf find}.\\ \midrule

\begin{tabular}[c]{@{}l@{}}{\tt rnr}: right node\\raising\end{tabular} & \begin{tabular}[c]{@{}l@{}}({\color{red}A}) verb/prep2\\({\color{blue}B}) verb/prep1\end{tabular} &
\begin{dependency}[edge above, edge slant=0.15ex, edge unit distance=0.6ex]
  \begin{deptext}[column sep=1ex]
    Humboldt \& supported \& and \& worked with \& other \& scientists.\\
  \end{deptext}
  \depedge{2}{6}{ARG2({\color{blue}B})}
  \depedge{4}{6}{ARG2({\color{red}A})}
\end{dependency} \\ \midrule
%Humboldt {\bf supported} and {\bf worked with} other {\it scientists}. \\ \midrule

\begin{tabular}[c]{@{}l@{}}{\tt absol}: absolutives\end{tabular} & \begin{tabular}[c]{@{}l@{}}({\color{red}A}) absolutive predicate\\({\color{blue}B}) main clause predicate\end{tabular} &
\begin{dependency}[edge above, edge slant=0.15ex, edge unit distance=0.6ex]
  \begin{deptext}[column sep=1ex]
    It \& consisted of \& 4 games \& each team \& facing \& other teams \& twice.\\
  \end{deptext}
  \depedge{5}{4}{ARG1({\color{red}A})}
  \depedge{2}{5}{MOD({\color{blue}B})}
\end{dependency} \\ \midrule
%Only spanish root words are {\bf listed} {\it derivations} not {\it \textbf{being included}}.\\ \midrule

\begin{tabular}[c]{@{}l@{}}{\tt vger}: verbal gerunds\end{tabular} &  \begin{tabular}[c]{@{}l@{}}({\color{red}A}) selecting head\\({\color{blue}B}) gerund\end{tabular} &
\begin{dependency}[edge above, edge slant=0.15ex, edge unit distance=0.6ex]
  \begin{deptext}[column sep=1ex]
    Asking for \& the \& help \& from the school \& prompts \& an \& announcement.\\
  \end{deptext}
  \depedge{5}{1}{ARG1({\color{red}A})}
  \depedge{1}{3}{ARG2({\color{blue}B})}
\end{dependency} \\ \midrule
%{\it Asking} for the ses {\bf prompts} a recorded announcement explaining how to proceed.\\ \midrule

\begin{tabular}[c]{@{}l@{}}{\tt control}: raising/\\control constructions\end{tabular} &  \begin{tabular}[c]{@{}l@{}}({\color{red}A}) ``upstairs'' verb\\({\color{blue}B}) ``downstairs'' verb\end{tabular} &
\begin{dependency}[edge above, edge slant=0.15ex, edge unit distance=0.6ex]
  \begin{deptext}[column sep=1ex]
    They \& managed \& to \& house \& and \& feed \& the \& poor.\\
  \end{deptext}
  \depedge{2}{1}{ARG1({\color{blue}B})}
  \depedge{2}{4}{ARG2({\color{red}A})}
  \depedge{2}{6}{ARG2({\color{red}A})}
\end{dependency} \\ \bottomrule
%{\it They} {\bf managed} to {\bf beat} USC during one of the few rainy UCLA-USC rivalry games.\\ \midrule
\end{tabular}}
\end{table*}

In this section, we discuss several important linguistic phenomena for evaluating semantic parsers, including lexical constructions,
predicate--argument structures, phrase constructions, and non-local dependencies \citep{fillmore1988regularity,kay1999grammatical,michaelis1996toward,hilpert2014construction}, which diverge from the common average-case evaluation but critical for understanding natural language \citep{goldberg2003constructions}. 
The phenomena and the corresponding examples are summarized in Table \ref{tb:linguistic phenomena}.

\subsection{Lexical Constructions: Multiword Expression}
Multiword Expressions (MWEs) are lexical items made up of multiple lexemes that undergo idiosyncratic constraints and therefore offer a certain degree of idiomaticity. 
MWEs cover a wide range of linguistic phenomena, including fixed and semi-fixed expressions, phrasal verbs, named entities.

Although MWEs can lead to various categorization schemes and its definitions observed in the literature tend to stress different aspects, 
in this paper we mainly focus on \textbf{compound} and \textbf{multiword named entity}. 
Roughly speaking, a compound is a lexeme formed by the juxtaposition of adjacent lexemes. 
Compounds can be subdivided according to their syntactic function. 
Thus, nominal compounds are headed by a noun (e.g., {\it bank robbery}) or a nominalized verb (e.g., {\it cigarette smoking}) 
and verbal compounds are head by a verb (e.g. {\it London-based}).
Multiword named entity is a multiword linguistic expression that rigidly designates an entity in the world, 
typically including persons, organizations, and locations (e.g., {\it International Business Machines}).

\subsection{Basic Argument Structure}
The term ``argument structure'' refers to a relationship that holds between a predicate denoting an activity, state, or event and the respective participants, which are called arguments.
Argument structure is often referred to as valency \citep{tesniere2018elements}. 
A verb can attract a certain number of arguments, just as an atom's valency determines the number of bonds it can engage in \citep{agel2009dependency}. 

Valency is first and foremost a characteristic of verbs, but the concept can also be applied to adjectives and nouns. For instance, the adjective {\it certain} can form a bond with a \textit{that}-clause in the sentence {\it I am certain that he left} or an infinitival clause ({\it John is certain to win the election}). Nouns such as {\it fact} can bond to \textit{that}-clause as well: \textit{the fact that he left}. We view all these valency relations as basic argument structures.

\subsection{Phrasal Constructions}
\label{sec:phrasal constructions}
In the past two decades, the constructivist perspective to syntax is more and more popular.
For example, \citet{goldberg1995constructions} argued that argument structure could not be wholly explained in terms of lexical entries alone, and syntactic constructions also lead hearers to understand some meanings.
Though this perspective is very controversial, we think the related phenomena are relevant to computational semantics. 
To test a parser's {\it adaptation} ability to handle {\it peripheral phenomena}, we evaluate the performance on several valency-increasing and decreasing constructions, 
including the ditransitive construction, cause motion construction, way construction and passive.

\paragraph{Ditransitive construction} The ditransitive construction links a verb with three arguments --- a subject and two objects. 
Whereas English verbs like {\it give}, {\it send}, {\it offer} conventionally include two objects in their argument structure, 
the same cannot be said of other verbs that occur with the ditransitive construction. 
\begin{itemize}
  \item[(1)] \it Sally baked her sister a cake.
\end{itemize}
The above sentence means that Sally produced a cake so that her sister could willingly receive it. 
We can posit the ditransitive construction as a symbolic unit that carries meaning and that is responsible for the observed increase in the valency of {\it bake}. In general, the ditransitive construction conveys, as its basic sense, the meaning of a transfer between an intentional agent and a willing recipient (indirect object).

\paragraph{Caused-motion construction} The caused-motion construction can also change the number of arguments with which a verb combines and yield an additional meaning. 
For example, in the sentence (2), the event structure of {\it laugh} specifies someone who is laughing and the unusually added argument leads to a new motion event in which both the agent and the goal are specified. 
\begin{itemize}
  \item[(2)] \it The audience laughed Bob off the stage.
\end{itemize}

\paragraph{Way construction} This construction specifies the lexical element {\it way} and a possessive determiner in its form. For example, in (3), the construction evokes a scenario in which an agent moves along a path that is difficult to navigate, thus adds up two arguments in the process --- the {\it way} argument and a path/goal argument that is different from the caused-motion construction.
\begin{itemize}
  \item[(3)] \it Frank dug his way out of prison.
\end{itemize}

\paragraph{Passive} The passive construction with {\it be} is most often discussed as the marked counterpart of active sentences with transitive verbs. For example, in (4) the subject of the active ({\it the reviewer}) appears in the corresponding passive sentences as an oblique object marked with the preposition {\it by}, and it is possible to omit this argument in the passive. It is this type of omission that justifies categorizing the passive as a valency-decreasing construction.
\begin{itemize}
  \item[(4)] \it The paper was accepted (by the reviewer).
\end{itemize}

\subsection{BFOZ's Ten Constructions}
\citet{bender-EtAl:2011:EMNLP} proposed a selection of ten challenging linguistic phenomena, each of which consists of 100 examples from English Wikipedia and occurs with reasonably high frequency in running text. 
Their selection (hereafter BFOZ) considers lexical (e.g., {\tt vpart}), phrasal (e.g., {\tt ned}) as well as non-local (e.g., {\tt rnr}) constructions.
The definitions and examples of the linguistic phenomena are outlined in Table \ref{tb:linguistic phenomena}, which considers representative local and non-local dependencies.
Refer to their paper for more information.

In some phenomena, there are subtypes A and B, corresponding to different arcs in the structure. 
It is noted that the number of ``A'' and the number of ``B'' are not necessarily equal, as illustrated in the example of {\tt control} in the table.
Some sentences contain more than one instance of the phenomenon they illustrate and multiple sets of dependencies are recorded. 
In total, the evaluation data consists of 2,127 dependency triples for the 1,000 sentences.

\section{Evaluation and Analysis}
To investigate the type of representations manipulated by different parsers, in this section, we evaluate the ACE parser and our parser regarding the linguistic phenomena discussed in Section \ref{sec:linguistic phenomena}.
In order to get the bi-lexical relationship, we use \textsc{smatch} to get all alignments between the output and ground truth.

\subsection{Lexical Construction}
\label{sec:lexical construction}
{\tt MRS} uses the abstract predicate {\tt compound} to denote compounds as well as light-weight named entities. The edge labeled with {\it ARG1} denotes the root of the compound structure and thus can help to distinguish the type of the compound (nominal or verbal compounds), and the nodes in named entities are labeled as {\it named-relation}. The head words of the compound in the test set can be other types such as adjectives, and due to their data sparsity in the test data, we just omit this part. The results are illustrated in Table \ref{tb:lexical construction}.

\begin{table}[H]
\caption{Accuracy of lexical constructions.}
\scalebox{1}{
\begin{tabular}{llrrrrrr}
\toprule
Type & Example & \# & ACE & W2V & ELMo & ELMo*\\\midrule
Compound  & -   & 2266 & 80.58 & 87.56 & 84.33 & \textbf{89.67} \\
\begin{tabular}[c]{@{}l@{}}Nominal {\it \scriptsize w/ nominalization}\end{tabular}& flag burning & 22  & 85.71  & \textbf{90.91} & 81.82 & \textbf{90.91} \\
\begin{tabular}[c]{@{}l@{}}Nominal {\it \scriptsize w/ noun}\end{tabular}& pilot union  & 1044 & 85.28 & 89.27 & 88.51 & \textbf{90.04} \\
Verbal                      & state-owned  & 23 & 40.91 & \textbf{82.61} & 47.83 & 65.22 \\
Named Entity                & Donald Trump & 1153 & 82.92  & 86.93 & 82.74 & \textbf{90.55} \\
\bottomrule
\end{tabular}}
\label{tb:lexical construction}
\end{table}

From the table, we find that ELMo* performs much better than ACE, especially for the named entity recognition (the total number of verbal compounds in the test set is rather small and does not affect the overall performance too much). It is noted that even the pure ELMo alone can achieve fairly good results, indicating that those pre-trained embedding-based models are good at capturing local semantic information such as compound constructions and named entities. 

\subsection{Basic Argument Structure}

The detailed performances on the 1474 test data in terms of basic argument structure are shown in Table \ref{tb:basic argument structure}. In {\tt MRS}, different senses of a predicate are distinguished by optional sense labels. Usually, the verb with its basic sense will be assigned the sense label as {\tt \_v\_1} (e.g., {\tt \_look\_v\_1}), while verb particle construction is handled semantically by having the verb contribute a relation particular to the combination (e.g., {\tt \_look\_v\_up}). In the evaluation, we also categorize the verb into basic verbs and verb particle constructions and show the detailed performance.

\begin{table}[ht]
\caption{Accuracies on basic argument structure over the 1474 test data. The accuracies are based on complete match, i.e., the predicates, arguments (nodes, edges and the edge labels in the graph) should be all correctly parsed to their gold standard graphs.}
\label{tb:basic argument structure}
\vspace{1em}
\scalebox{1}{
\begin{tabular}{l|rrrrr}
\toprule
Type & \# & ACE & W2V & ELMo & ELMo*\\\midrule
Overall & 7108 & \textbf{86.98} & 81.44 & 74.56 & 84.70\\\midrule
Total verb & 4176 & \textbf{85.34} & 77.59 & 69.08 & 81.81\\\midrule
Basic verb & 2354 & \textbf{85.79} & 80.76 & 73.70 & 83.90 \\
\ \ \textit{ARG1} & 1683 & \textbf{90.25} & 87.17 & 80.45 & 89.07 \\
\ \ \textit{ARG2} & 1995 & \textbf{90.48} & 84.96 & 81.95 & 87.85 \\
\ \ \textit{ARG3} & 195  & \textbf{82.63} & 58.46 & 55.90 & 72.31 \\\midrule
Verb-particle & 1761 & \textbf{84.69} & 73.31 & 62.86 & 78.99 \\
\ \ \textit{ARG1} & 1545 & \textbf{89.57} & 80.45 & 75.15 & 84.72 \\
\ \ \textit{ARG2} & 923  & \textbf{86.27} & 78.80 & 68.42 & 82.73 \\
\ \ \textit{ARG3} & 122  & \textbf{81.88} & 58.44 & 47.40 & 73.38 \\\midrule
Total noun & 394 & \textbf{92.41} & 87.56 & 72.34 & 90.61 \\
Total adjective & 2538 & \textbf{89.27} & 87.31 & 84.48 & 89.05 \\
\bottomrule
\end{tabular}}
\end{table}

As can be observed from the table, the overall performance of ELMo* is relatively worse than the one of ACE, and this is mainly due to the relatively low accuracy of verb particle constructions and ARG3. As for the pure ELMo model, this issue will be exacerbated. The verb particle construction emphasizes combinations, and ARG3 often denotes long distances cross words within the sentence, while pure ELMo (without LSTM) is weak in capturing such information.

\subsection{Phrasal Construction}
For each valency increasing construction (ditransitive construction, caused-motion construction, and way construction) introduced in Section \ref{sec:phrasal constructions}, based on some pre-defined patterns, we first automatically got a large scale of candidate sentences from Linguee\footnote{\url{https://www.linguee.com/}}, a web service that provides access to large amounts of appropriate bilingual sentence pairs found online. The paired sentences identified undergo automatic quality evaluation by a human-trained machine learning algorithm that estimates the quality of those sentences. We manually select 100 sentences for each type of linguistic phenomena to guarantee the quality of pattern-based selection in terms of correctness and diversity. In order to form the gold standard for the subsequent evaluation, we then ask a senior linguistic student who is familiar with {\tt ERG} to annotate the argument structure of those sentences. The annotation format is based on dependency triples, identifying the head words and dependents by the surface form of the head words in the sentence suffixed with a number indicating the word position.

As for the valency decreasing construction, viz. the passive construction, {\tt MRS} gives special treatment to passive verbs, identified by the abstract node {\tt parg\_d}. Similar to the previous evaluation, we test the parsing accuracy on {\tt parg\_d} over the 1474 test data. The results of phrasal constructions are shown in Table \ref{tb:valency evaluation}.

\begin{table}[H]
\caption{Accuracies on phrasal construction, including the valency increasing evaluation over 300 selected sentences and the valency decreasing evaluation over the 1474 test data.}
\label{tb:valency evaluation}
\vspace{1em}
\scalebox{1}{
\begin{tabular}{l|rrrrr}
\toprule
Type & \# & ACE & W2V & ELMo & ELMo*\\\midrule
\texttt{ditr}  & 100 & 87.36 & 90.00 & 88.00 & \textbf{93.00}\\
\ \ \textit{ARG1} & 98  & \textbf{97.65} & 95.92 & 94.90 & 96.94\\
\ \ \textit{ARG2} & 100 & \textbf{100.00} & 99.00 & 98.00 & 99.00\\
\ \ \textit{ARG3} & 100 & 87.36 & 94.00 & 93.00 & \textbf{95.00} \\\midrule
\texttt{causemo}  & 100 & 41.11 & 27.00 & 32.00 & \textbf{55.00} \\
\ \ \textit{ARG1} & 94  & 91.86 & 90.43 & 75.53 & \textbf{93.62}\\
\ \ \textit{ARG2} & 100 & \textbf{100.00} & 99.00 & 97.00 & 99.00\\
\ \ \textit{ARG3} & 100 & 43.33 & 30.00 & 45.00 & \textbf{60.00}\\\midrule
\texttt{way}      & 100 & \textbf{7.14} & 0.00 & 3.00 & 4.00\\
\ \ \textit{ARG1} & 94  & 81.25 & 86.46 & 79.17 & \textbf{88.54}\\
\ \ \textit{ARG2} & 100 & 61.22 & 96.00 & 59.00 & \textbf{99.00}\\
\ \ \textit{ARG3} & 100 & \textbf{9.18} & 1.00 & 4.00 & 7.00\\\midrule
\texttt{passive} & 522 & \textbf{85.12} & 82.57 & 76.05 & 84.87 \\
\bottomrule
\end{tabular}}
\end{table}

The results are shown in Table \ref{tb:valency evaluation}, from which we find that all the parsers perform worse on the way construction, while on the other valency increasing constructions, ELMo* yields the best results. The performances on the three constructions are mainly affected by the performances on {\it ARG3}, where ELMo* performs relatively better on ditransitive and caused-motion constructions. Interestingly, it is a contrast to the results on {\it ARG3} in basic argument constructions.

The parsers may run across a variety of cases where a predicate appeared to be in an atypical context: none of the senses listed in the lexicon provided the appropriate role label choices for novel arguments encountered. The issue can be addressed by either adding many individual senses for every predicate compatible with a particular construction, as what rule-based parser has done, or learning the construction pattern from the training data, as what data-driven parser has done. 

The latter option clearly had a practical advantage of requiring far less time-consuming manual expansion of the lexicon while may also suffering from data sparsity. The annotated {\tt MRS} data provides considerably wide coverage for the most frequent and predictably patterned linguistic phenomena, while sometimes fails to include some of the rarer structures found in the long tail of language. According to our statistics, the cause motion and way constructions are very sparse in the training data -- appearing 12 and 0 times respectively in the 35,314 sentences, severely limiting the prediction on these constructions.

% with no training data
% with no pre-defined lexical unit

\begin{table}[H]
\caption{Recall of individual dependencies on \citeauthor{bender-EtAl:2011:EMNLP}'s ten constructions. Arc refers to the average distance between the two nodes; $\Delta_{+x}$ refers to the improvement of performance when add feature $x$, compared with random.}
\scalebox{0.88}{
\begin{tabular}{l|r|rrrrrr|rrr}
\toprule
Phenomena & Arc & ACE$_1$ & AEC$_2$  & Rand & W2V & ELMo & ELMo* & $\Delta_{\text{\tiny{+W2V}}}$ & $\Delta_{\text{\tiny{+ELMo}}}$ & $\Delta_{\text{\tiny{+LSTM}}}$ \\\midrule
{\tt vpart} & 3.8 & 79 & 81 & 71 & 69 & 46 & \textbf{85} & -2 & +14 & +39 \\
\midrule
{\tt itexpl} & - & \textbf{91} & \textbf{91} & 52 & 48 & 63 & 74 & -4 & +23 & +11 \\
{\tt ned(A)} & 2.7 & 63 & 72 & 78 & 83 & 79 & \textbf{88} & +5 & +10 & +9 \\
{\tt ned(B)} & 1 & 81 & \textbf{93} & 75 & 79 & 47 & 83 & +4 & +7 & +35 \\
{\tt argadj(A)} & 1.6 & 78 & \textbf{84} & 74 & 75 & 69 & 76 & +1 & +3 & +8 \\
{\tt argadj(B)} & 6.3 & 50 & 53 & 39 & 47 & 43 & \textbf{56} & +8 & +17 & +13 \\
\midrule
{\tt barerel} & 3.4 & 60 & 67 & 70 & 72 & 73 & \textbf{75} & +2 & +6 & +3 \\
{\tt tough(A)} & 2.2 & 88 & 90 & \textbf{91} & 90 & 90 & 86 & -1 & -5 & -4 \\
{\tt tough(B)} & 6.4 & 83 & \textbf{85} & 68 & 70 & 47 & 83 & +2 & +16 & +37 \\
{\tt rnr(A)} & 2.8 & 69 & 76 & 75 & 72 & \textbf{77} & 73 & -3 & -2 & -4 \\
{\tt rnr(B)} & 6.2 & 43 & \textbf{47} & 17 & 17 & 10 & 32 & +1 & +16 & +22 \\
{\tt absol(A)} & 1.8 & 61 & 68 & 81 & 83 & 73 & \textbf{92} & +3 & +11 & +18 \\
{\tt absol(B)} & 9.5 & 6 & \textbf{7} & 3 & 3 & 3 & 3 & 0 & 0 & 0 \\
{\tt vger(A)} & 1.9 & 56 & 62 & \textbf{69} & 62 & 61 & \textbf{69} & -7 & 0 & +8 \\
{\tt vger(B)} & 2.4 & 80 & 88 & 88 & \textbf{89} & 79 & 84 & +2 & -4 & +5 \\
{\tt control(A)} & 3 & 90 & 91 & 83 & 87 & 82 & \textbf{92} & +4 & +8 & +9 \\
{\tt control(B)} & 4.8 & 87 & \textbf{89} & \textbf{89} & 88 & 63 & 91 & -1 & +2 & +28 \\
\bottomrule
\end{tabular}}
\label{tb: non-local dependecies}
\end{table}

\subsection{BFOZ's Ten Constructions}

While the annotated style for those 10 linguistic phenomena introduced in \citet{bender-EtAl:2011:EMNLP} is not the same as the one of {\tt MRS}, we were able to associate our parser-specific results with the manually-annotated target non-local dependencies, and Table \ref{tb: non-local dependecies} shows the results.

All the parser perform markedly worse on the dependencies of {\tt rnr(B)}, {\tt absol(B)} and {\tt argadj(B)}, which have very long average distances of dependencies. 
Each of the parsers attempts with some success to analyze each of these phenomena, but they vary across phenomena:

\begin{enumerate}
    \item Comparing pure ELMO and ELMO*, we can observe that in most cases, ELMO* outperforms pure ELMO especially for long-distance dependencies such as {\tt tough(B)}, {\tt vpart} and {\tt control(B)}, indicating that LSTM features helps to capture distant information to some extent. For example, {\tt tough(B)} is a relatively long-distance relation, and the significant improvement could be observed when we add ELMo and LSTM.
    
    \item Similar to the conclusion drawn in \ref{sec:lexical construction}, in general, compared with ACE, ELMO is good at capturing relatively local dependencies that have short distances, e.g., {\tt absol(A)}, {\tt vger(A)} and {\tt ned(A)}.
    
    \item For some of the phenomena, such as {\tt itexpl}, ACE works pretty well while all the neural models fail to achieve competitive results, and this is because {\tt itexpl} could be completely covered by pre-define grammar while it is very hard to learn the relation implicitly. On the other hand, the knowledge-intensive parser will be confused when handling phenomena that are not covered by pre-defined grammar, e.g., {\tt barerel}, {\tt absol(A)}.
\end{enumerate}

\begin{figure}
    \caption{Performances of compound (compound), named entity (ner) and basic argument (arg) on development set when down-sampling the training data size}
    \label{fig:down sample1}
    \vspace{1em}
    \begin{minipage}[t]{0.72\textwidth}
        \centering
        \includegraphics[width=\textwidth]{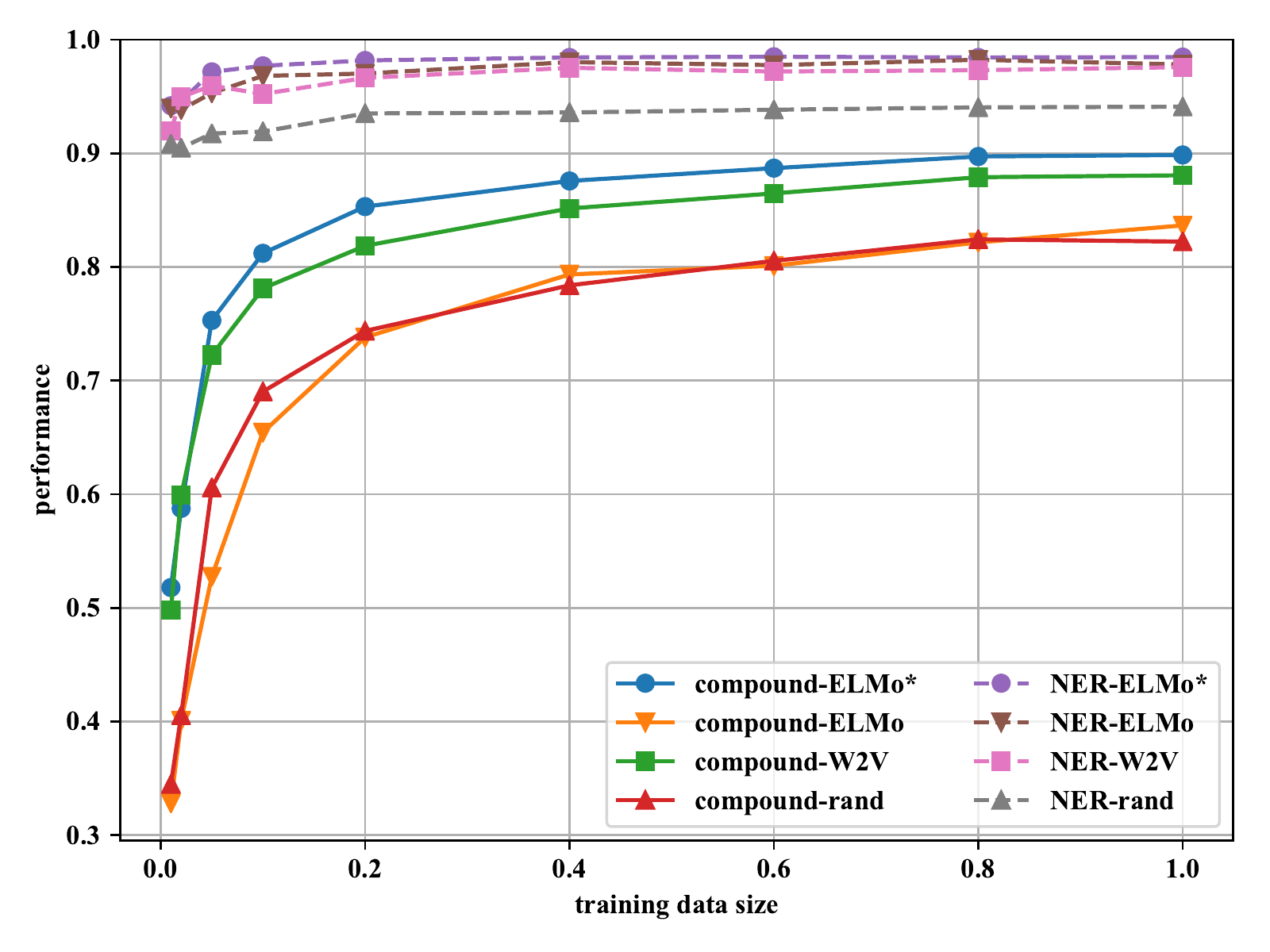}
        %\caption{Lexical constructions}
    \end{minipage}%
    \\
    \begin{minipage}[t]{0.72\textwidth}
        \centering
        \includegraphics[width=\textwidth]{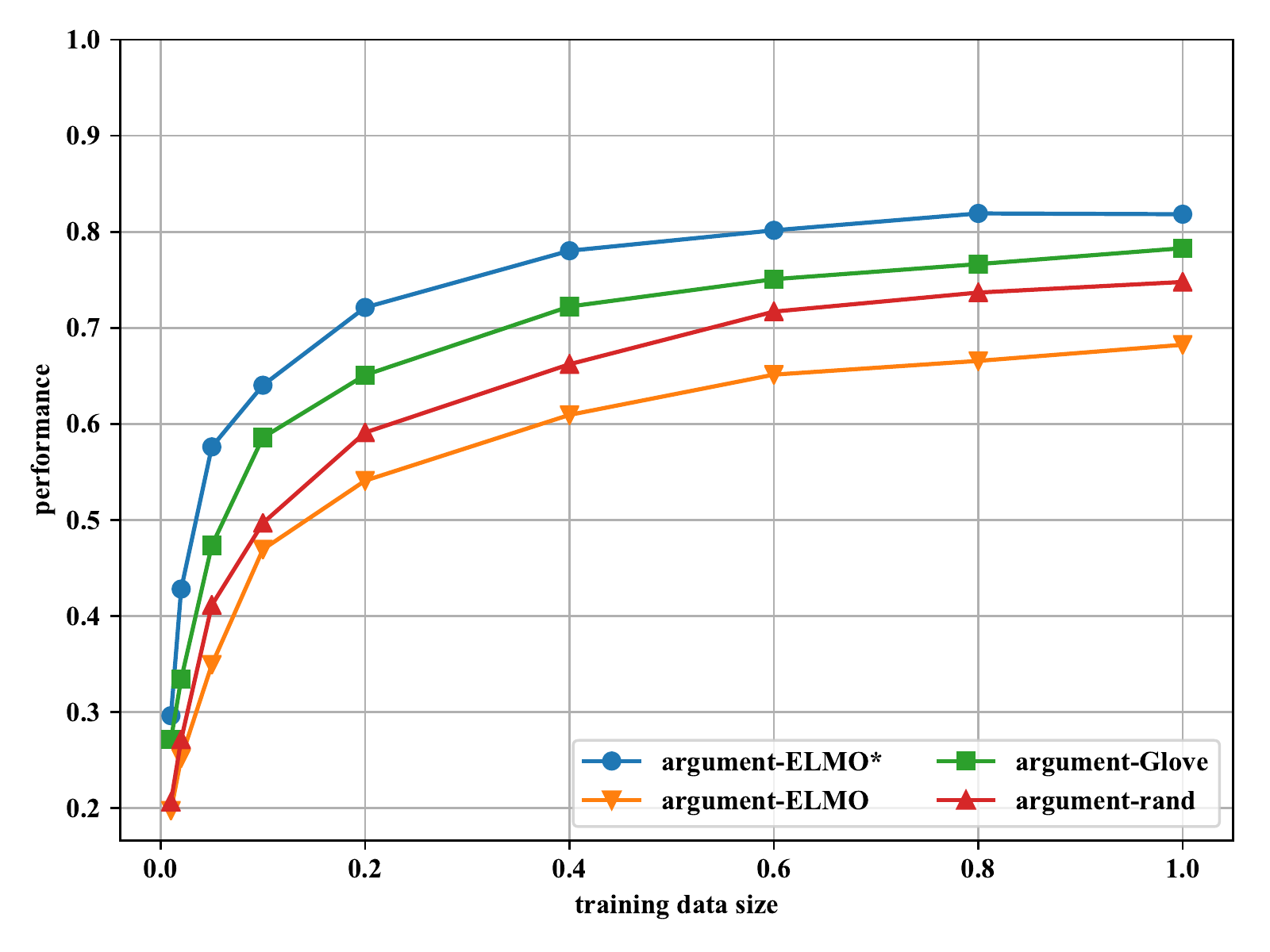}
        %\caption{Argument constructions}
    \end{minipage}
\end{figure}

\begin{figure}
  \caption{Performances of valency-increasing constructions (valency) and passive (passive) on development set when down-sampling the training data size}
  \label{fig:down sample2}
    %\begin{minipage}[t]{0.48\textwidth}
        \includegraphics[width=0.72\textwidth]{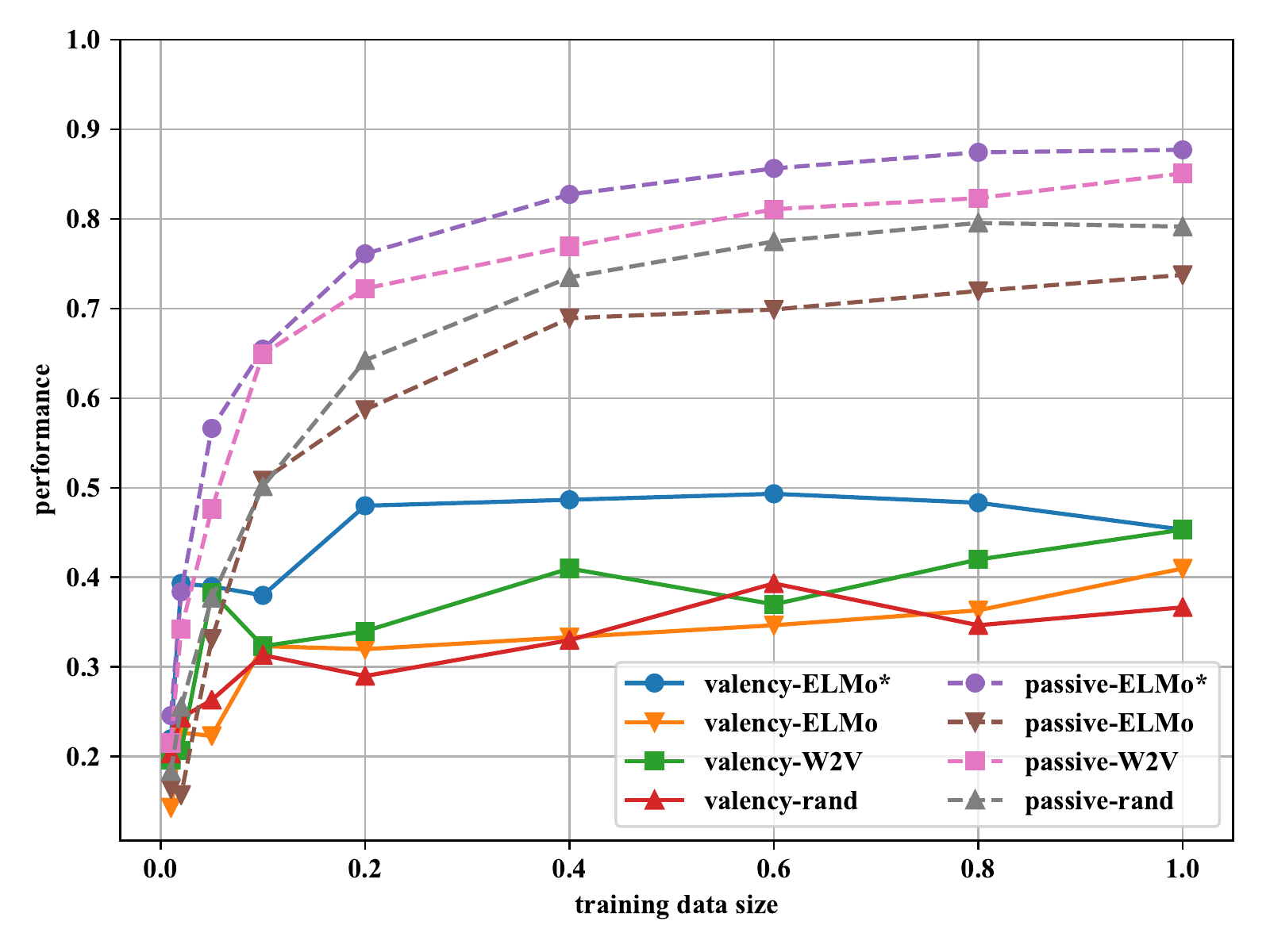}
        %\caption{Phrasal constructions}
    %\end{minipage}%
\end{figure}

\subsection{Down-sampling Data Size}

Our next experiment examines this effect in a more controlled environment by down-sampling the training data and observing the performances on the development set. The results are shown in Figure \ref{fig:down sample1} and \ref{fig:down sample2}, where we test the overall performance for lexical (compound and named entities), basic argument and phrasal (valency-increasing and passive) constructions.

As can be seen from Figure \ref{fig:down sample1}, adding training data cannot help the parser predict valency-increasing constructions that much. When it comes to local constructions (lexical construction), even rather small training data can lead to relatively high performance, especially for the light-weight named entity recognition. The learning curve of the basic argument structures also serves as another complementary reflection. From Figure \ref{fig:down sample2}, we also find that due to the low frequency of the valency-increasing constructions in the data, the performance will stay low as the training data grows.

\section{Conclusion and Discussions}

In this work, we proposed a new target structure-centric parser that can produce semantic graphs (\eds~here) much more accurately than previous data-driven parsers.
Specifically, we achieves an accuracy of 95.05 for Elementary Dependency Structure (\eds) in terms of \textsc{smatch}, 
which yields a significant improvement over previous data-driven parsing models. 
Comparing this data-intensive parser to the knowledge-intensive ACE parser sheds light on complementary strengths different type of parser exhibits.
%  which is 8.05 point improvement over the best transition-based model \citep{buys-blunsom:2017:Long} 
%  and 4.19 point improvement over the composition-based parser \citep{shrgparser}. 
  %and 0.45 point improvement over the knowledge-intensive parser ACE.

%To deeply understand the difference between knowledge-intensive and data-intensive parsing technologies, 
To this end, we have presented a thorough study of the difference in errors made between systems that leverage different methods to express the mapping between string and meaning representation.
To achieve that, we employed a construction-focused parser evaluation methodology as an alternative to the exclusive focus on incremental improvements in overall accuracy measures such as \textsc{smatch}.
We have shown that knowledge- and data-intensive make different types of errors and such differences can be quantified concerning linguistic constructions.
Our analysis provides insights that may lead to better semantic parsing models in the future.
Below we sketch some possible new directions:
\begin{itemize}
\item neural tagging for the knowledge-intensive parser, 
\item structural inference for the data-intensive parser,
\item syntactic knowledge for the data-intensive parser, and
\item parser ensemble.
\end{itemize}

%\starttwocolumn
\bibliographystyle{compling}
\bibliography{compling_style}

\end{document}